# Minimal Assumption Distribution Propagation in Belief Networks

**Ron Musick**
Computer Science Division
University of California
Berkeley, CA 94720
musick@cs.berkeley.edu

## Abstract

As belief networks are used to model increasingly complex situations, the need to automatically construct them from large databases will become paramount. This paper concentrates on solving a part of the belief network induction problem: that of learning the quantitative structure (the conditional probabilities), given the qualitative structure. In particular, a theory is presented that shows how to propagate inference distributions in a belief network, with the only assumption being that the given qualitative structure is correct. Most inference algorithms must make at least this assumption. The theory is based on four network transformations that are sufficient for any inference in a belief network. Furthermore, the claim is made that contrary to popular belief, error will not necessarily grow as the inference chain grows. Instead, for QBN belief nets induced from large enough samples, the error is more likely to decrease as the size of the inference chain increases.

## 1 INTRODUCTION

The area of belief networks (and more generally influence diagrams) is one of the fastest growing disciplines in AI, because they provide a principled yet efficient manner with which to apply probability theory to the problems of reasoning, modeling and decision making. As the belief network technology is applied to more diverse and complicated scenarios, the ability to automatically induce part, or all of a belief network will become essential. The problem is that as the complexity of the situations being modeled grows, both the number of possible network structures and the number of conditional probabilities in each structure explode. The larger the human specified portion of the network, the larger the chance for errors, and the more the overall construction process will cost. In addition, as humans our ability to make qualitative judgements about which variables affect the value or state of another is good; however, our ability to make quantitative statements is extremely poor (Wexelblat, 1989). For example, where we might be able to say that weather has an effect on traffic, when asked to predict numerically how the throughput on I880 will vary with snow, we probably would be unable to make an accurate guess.

This paper focuses on the weak link in the chain: the induction of the conditional probabilities, given the graphical network structure (the qualitative structure). A straightforward approach to learning conditional probabilities of variables in a database is to statistically infer them by sampling (Piatetsky-Shapiro, 1991; Cooper and Herskovits, 1992)[1]. These values can be learned as point probabilities, intervals (Fertig and Breese, 1990), or even as second order distributions (Klieter, 1992; Musick *et al.*, 1993; Spiegelhalter and Lauritzen, 1990). Distributions are generally preferable to point probabilities or intervals because they contain more information, and thus are more useful. For example, distributions give the ability to assign some degree of confidence to the inferences performed; they also allow comparison between inference solutions produced by different networks or algorithms. There is potential for much more. To date, however, there is no completely satisfactory method with the capability of propagating distributions correctly through inference; the assumptions that have been required compromise the validity or generality of the resulting learned distributions.

This paper provides a theory based on a set of four network transformations that can produce any inference distribution in a given network, with the only assumption being that the given qualitative structure of the network is correct. This is really a proof of concept that inference distributions can be propagated without significant loss, and without debilitating assumptions. This theory will be applied in QBN (Quantified Belief Network), a testbed for research on learning quantitative structures of belief networks. QBN belief net-works have distributions stored at every cell in the conditional probability tables, instead of point probabilities. The distributions can be manipulated with the same ease as point probabilities (computationally), with the distinct advantage of providing all the information necessary to make judgements on the precision of the sampled probabilities. The QBN makes other distinctions as well, but they are beyond the scope of this paper.

[1] Cooper and Herskovits actually use the entire database, treating it as one large sample.

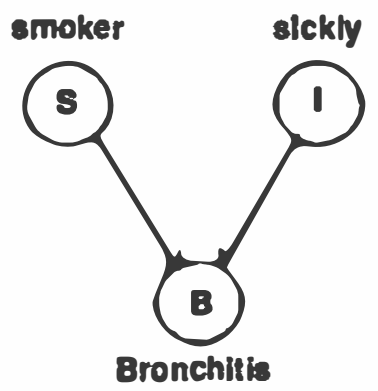

Figure 1: **A Simple Belief Net**
This shows a model of the dependency of bronchitis on whether the patient is sickly, and whether the patient smokes.

In Section 2 previous work in this area is examined. Section 3 establishes the framework for the induction of the quantitative structure of belief nets. In Section 4, a sufficient set of transformations for any inference in a belief network is shown, and proofs are given that show how the inference distributions are correctly propagated through the transformations. Section 5 develops an example that is meant to clarify the concepts presented in the paper. Section 6 is the conclusion.

## 2 PREVIOUS WORK

The idea of sampling to build second order distributions is not new; most of the work discussed in this section has used some form of a distribution to represent the conditional probabilities in the network. The difference comes when we consider how the distributions are used for inference.

The approach taken by Spiegelhalter and Lauritzen (Spiegelhalter and Lauritzen, 1990; Spiegelhalter *et al.*, 1992) is to specify two types of variables in the network: *core* variables, and *uncertainty* variables. The core variables correspond to typical belief net variables. They have a set of core variable parents, and a set of uncertainty variable parents (one for each unique instantiation of the core variable parents). The uncertainty variables provide a discrete specification of the second order distribution of the corresponding conditional probabilities. Spiegelhalter and Lauritzen show how to update these uncertainty variables, and mention their use in providing estimates of the inference distributions as inference is done. The selling point of this approach is the smooth integration of the uncertainty variables into a speedy inference algorithm. On the down side, there are two factors that degrade the quality of these distributions. First, a strong independence assumption is made (termed local independence). The assumption is that the uncertainty variables attached to a particular core variable are mutually independent. As an example of what this implies, from Figure 1 we could say that the distribution of $Pr(\textbf{bronchitis} \mid \textbf{sickly}, \textbf{smokes})$ is independent of the distribution of $Pr(\textbf{bronchitis} \mid \overline{\textbf{sickly}}, \textbf{smokes})$. The second factor is that the uncertainty variables must approximate continuous distribution functions with discrete representations.

Klieter (Klieter, 1992) starts in a similar fashion to Spiegelhalter and Lauritzen by transforming human supplied data into corresponding beta distributions. Then, for diagnostic inferences in a tree structured belief network (actually, an expert system), an expression for the distribution of the inference is produced. This expression is approximated as a beta distribution, where the mean and variance for the beta are produced from approximations of the mean and variance of the original expression. This approach, though complicated, yields good results for the particular inference problem that is analyzed. It would require more work to extend the results to inference within general DAG structured belief networks.

Fertig and Breese (Fertig and Breese, 1990) propose an approach based on interval probabilities. Using the fact that there is a set of network transformations with which any inference can be done in a influence diagram (a proof of this is alluded to in Olmstead's thesis and a sketch of it is given in Section 4), they show how to keep consistent upper and lower bounds on inference probabilities. Unfortunately, the bounds tend to get weaker with each transformation, and intervals are not as informative as actual distributions. The concept of there being a small set of transformations sufficient to describe any inference in a belief net is, however, central to this paper.

## 3 FRAMEWORK FOR LEARNING QBNS

We now turn to exploring the QBN approach. The purpose of this section is to clarify exactly what the distributions for the QBN are, where they come from, and how they will be learned from samples.

We are given a database of instances $D$, and a sample[2] $S \subseteq D$ drawn by sampling with replacement. Let $B_D$ be the belief net that corresponds to the underlying model from which $D$ was drawn. There are $n$ nodes $X_1, \ldots, X_n$ in $B_D$, where node $X_i$ takes values from the set $d_{i,l} \in \mathcal{D}_i$. A complete sample $s$ is an element of $S$ which assigns a value from $\mathcal{D}_i$ to every node $X_i$. Let $\prod_i$ be the parents of node $X_i$, $\phi_i$ be the set of unique instantiations of the parents $\prod_i$, and $\phi_i[j]$ be the jth unique instantiation. Finally, $\iota_{i,j,k}$ is the combination of the parents instantiation $\phi_i[j]$ with the node instan-tiation $d_{i,k}$.

[2]The term sample is used to describe both a set of instances, and a single instance in this paper.

The formal update process, the mechanism by which the conditional probability tables will be learned, is derived from the following arguments. In the belief net $B_D$ there are a set of unknown conditional probabilities $\theta_{i,j,k}$ that we are attempting to estimate for each possible instantiation $\iota_{i,j,k}$ in the network. The estimation problem can be considered one of statistical inference in which observations have been taken from a p.d.f. $f(s_l|\theta_{i,j,k})$, where $\theta_{i,j,k}$ is unknown. Take $p$ independent random samples $s_1, \ldots, s_p$ from a distribution $f(s_l|\theta_{i,j,k})$. Let the joint p.d.f. of the $p$ samples be

$$\begin{aligned} f_p(\mathrm{s}|\theta_{i,j,k}) &= f_p(s_1, \ldots, s_p|\theta_{i,j,k}) \\ &= f(s_1|\theta_{i,j,k}) \cdots f(s_p|\theta_{i,j,k}). \end{aligned}$$

Choose some prior distribution $\xi(\theta_{i,j,k})$ for $\theta_{i,j,k}$. The posterior distribution $\xi(\theta_{i,j,k}|\mathrm{s})$, which is the estimate of $\theta_{i,j,k}$, is then found as

$$\xi(\theta_{i,j,k}|\mathrm{s}) = \frac{f_p(\mathrm{s}|\theta_{i,j,k})\xi(\theta_{i,j,k})}{\int_\Omega f_p(\mathrm{s}|\theta_{i,j,k})\xi(\theta_{i,j,k})d\theta_{i,j,k}} \quad \text{for } \theta_{i,j,k} \in \Omega,$$

which is proportional to $f_p(\mathrm{s}|\theta_{i,j,k})\xi(\theta_{i,j,k})$.

When sampling with replacement from the database $D$, a natural description of the sample distribution $f(s_l|\theta_{i,j,k})$ is as a Bernoulli distribution[3]; in a relevant sample, there is a $\theta_{i,j,k}$ chance that the sample will have $X_i$ assigned to $d_{i,k}$ (given that the parents $\prod_i$ have the assignment $\phi_i[j]$), and a $1 - \theta_{i,j,k}$ chance that $X_i$ will have a different value. With priors distributed as beta distributions, the posterior distributions will be betas as well. More precisely, if the prior is a beta with parameters $a$ and $b$ ($\beta(a, b)$), and we take $p$ samples, $y$ of which are successful (meaning $X_i = d_{i,k}, \prod_i = \phi_i[j]$), then the posterior is $\beta(a + y, b + p - y)$. A formal proof of this can be found in (DeGroot, 1986).

For learning QBNs, each conditional probability is represented by the sufficient statistics $\alpha$ and $\omega$ of a beta distribution. Though the prior in each cell of the table for node $X_i$ is arbitrarily set to $\beta(\alpha = 1, \omega = |\mathcal{D}_i| - 1)$, the theorems in the following section depend only on the prior being a beta. With the priors established, the induction of the quantitative structure of the network is simply a matter of incrementing the $\alpha$ and $\omega$ statistics of each conditional probability for each relevant sample seen. A sample is relevant to the conditional probability $Pr(X_i = d_{i,k}| \prod_i = \phi_i[j])$ if the sample is consistent with $\prod_i = \phi_i[j]$. The $\alpha$ statistic is incremented if both $X_i = d_{i,k}$ and $\prod_i = \phi_i[j]$ hold in the sample; the $\omega$ statistic is incremented if $X_i \neq d_{i,k}$, but $\prod_i = \phi_i[j]$.

[3]A sample distribution can be equivalently described as a multinomial over the joint space, in which case the priors and posteriors would be Dirichlet distributions.

## 4 PROPAGATING DISTRIBUTIONS

The estimate $B_S$ of $B_D$ that is being constructed from a sample $S$ has an updatable beta distribution for every conditional probability that is to be explicitly stored in the network. For very simple inference tasks, like those that ask for probabilities that are already explicit in the network, the mean and variance of the corresponding beta can be returned as the result. But when a more general inference is desired, distributions across nodes must be combined. This is a difficult problem because beta distributions are not conjugate across addition. Without strong assumptions about independence, or without approximation, there exists no readily apparent way to combine them.

The trick to combining the distributions is to glom the sample frequencies required to describe the new distribution from the $\alpha$ and $\omega$ statistics of the given distributions. This is not a general approach (again, betas are not conjugate across addition); there will be a unique way to obtain each desired frequency for each particular transformation. The approach is viable, however, because we need only find the betas for a few fundamental transformations.

Let $\Omega$ and $\Psi$ be mutually exclusive subsets of the nodes in the network, and let $X_{il}$ be an instantiation of $X_i \in \Omega$ with value $d_{i,l}$, and $X_{jk}$ be an instantiation of $X_j \in \Psi$. An inference problem can be defined as the task to provide a value for a query of the form:

$$Pr(\bigwedge X_{il} \mid \bigwedge X_{jk}). \tag{1}$$

It has been alluded to in (Olmstead, 1983) that there are a fundamental set of transformations on an influence diagram, and that all inferences can be done in terms of these transformations. The transformations are Node Removal, Arc Reversal, Node Merging and Node Splitting, which are shown in part in Figures 2, 3, 4 and 5. Olmstead uses these transformations to show how to propagate point probabilities through the network while doing inference. We use the same transformations in a belief network, demonstrating how to propagate the inference *distributions*, as opposed to the point probabilities.

Two main concepts form the basis of the proofs in this section. One is that the inference distribution is beta representable. From Section 3, it is clear that when considering one cell of a conditional probability table, $Pr(X_i = d_{i,k}| \prod_i = \phi_i[j])$, the $\alpha$ statistic represents all of the samples that fall into that cell (all samples consistent with $X_i = d_{i,k}, \prod_i = \phi_i[j]$), and the $\omega$ statistic represents the samples that satisfied the parent variable instantiations $\phi_i[j]$, but not the value $d_{i,k}$ for node $X_i$. When the distribution for $Pr(X_i = d_{i,k}| \prod_i = \phi_i[j])$ is considered in terms of the joint space over the network, it can be seen that $\alpha$ and $\omega$ are actually statistics over two clusters of cells in the joint space: the cluster of cells that correspond to the case where $X_i = d_{i,k}$ and $\prod_i = \phi_i[j]$, and the cluster of cells corresponding to $X_i \neq d_{i,k}$ and $\prod_i = \phi_i[j]$. A simple extension of this idea leads to the fact that any inference problem in the form of Equation 1 can be structured in terms of two clusters of cells over the joint space, and thus is beta representable.

The other commonality in the proofs is that they all require the Network Assumption.

**Network Assumption:** Assume that the independence conditions stated implicitly in the given qualitative structure are correct.

For the network fragments in Figures 2, 3, 4, and 5, this assumption allows the following simplification to take place:

$$
\begin{aligned}
&Pr(X_{il}, X_{jk}, V_{1p}, V_{2q}, V_{3r}) \\
&\quad = Pr(X_{il}, X_{jk}, V_{2q}, V_{3r}) Pr(V_{1p} | X_{il}, X_{jk}, V_{2q}, V_{3r}) \\
&\quad = Pr(X_{il}, X_{jk}, V_{2q}, V_{3r}) Pr(V_{1p} | X_{jk}, V_{2q}) \\
&\quad = \frac{Pr(X_{il}, X_{jk}, V_{2q}, V_{3r}) Pr(V_{1p}, X_{jk}, V_{2q})}{Pr(X_{jk}, V_{2q})}.
\end{aligned}
$$

The beta distribution representing the joint probability is based on frequencies derived from the samples. Let $\#X_{il}$ represent the number of samples in which variable $X_i$ takes the value $d_{i,l}$. Then

$$
\begin{aligned}
&Pr(X_{il}, X_{jk}, V_{1p}, V_{2q}, V_{3r}) \\
&\quad = \frac{\#X_{il} \wedge X_{jk} \wedge V_{1p} \wedge V_{2q} \wedge V_{3r}}{\text{Total } \# \text{ samples}} \\
&\quad = \frac{\frac{(\#X_{il} \wedge X_{jk} \wedge V_{2q} \wedge V_{3r})(\#V_{1p} \wedge X_{jk} \wedge V_{2q})}{\#X_{jk} \wedge V_{2q}}}{\text{Total } \# \text{ samples}}. \qquad (2)
\end{aligned}
$$

The rest of this section gives the proofs showing the network transformations and the corresponding inference distributions. At the end of the section an outline is given of the proof that the transformations are sufficient for any inference in a belief network. In the theorems below, in order to diminish the notational nightmare, spurious subscripts have been removed except where necessary. For example, the probability $Pr(X_{il} | X_{jk}, V_{2q}, V_{3r})$ might be represented as $\alpha_1$, whereas the complete description would be $\alpha_{1,l,k,q,r}$. They both refer to the alpha statistic in the cell for which $X_i = d_{i,l}, \ldots, V_3 = d_{3,r}$.

The Arc Reversal Theorem describes the arc reversal operation in full generality. The Node Removal, Node Splitting and Node Merging Theorems are a bit weaker than they could be, since the network fragments given are not completely general. This simplifies the presentation of the theorems without losing the ability to do every inference in the network.

Finally, the proofs below are constructed with $\beta(0,0)$ priors, which allow the theory to be used with arbitrary priors. To add an informed prior, the theorems can be strictly applied as written; the priors will simply be treated as samples already seen. For uninformed priors, the prior must be stripped from the distribution before the transformation is performed; after the transformation the new prior can be added back to the result. The uninformed prior is specially treated in order to prevent it from inadvertently providing information due to the transformations.

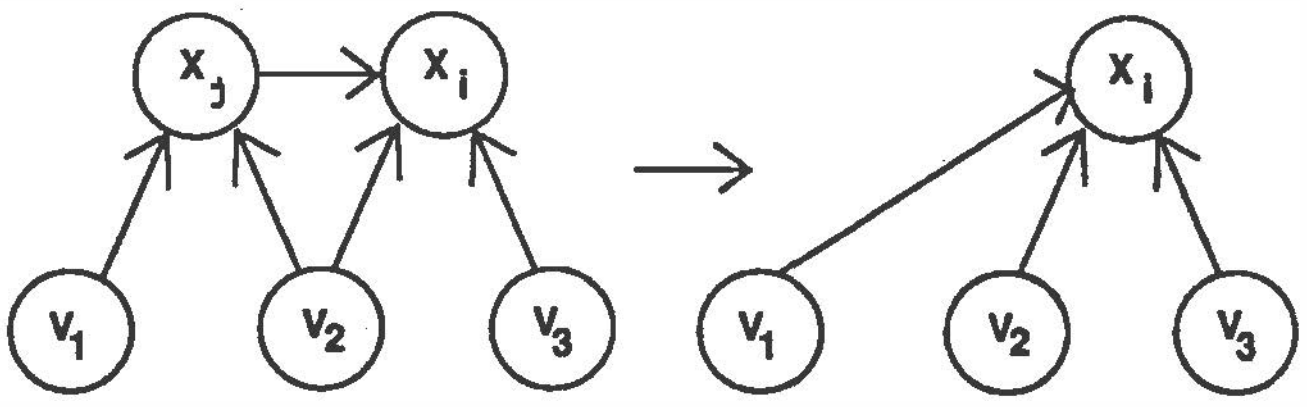

Figure 2: **Node Removal**

$V_1$ represents the set of nodes which are parents of $X_j$ but not $X_i$; $V_2$ represents the set of parents common to both $X_i$ and $X_j$; and $V_3$ is the set of nodes which are parents of $X_i$ but not $X_j$.

### Theorem 4.1 Node Removal

*Given:*
*The Network Assumption, and the network fragment as described in Figure 2, where $V_1$ represents the set of nodes which are parents of $X_j$ but not $X_i$; $V_2$ represents the set of parents common to both $X_i$ and $X_j$; and $V_3$ is the set of nodes which are parents of $X_i$ but not $X_j$. Also, given:*

$$
\begin{aligned}
&Pr(X_{il} | X_{jk}, V_{2q}, V_{3r}) \quad \textit{distributed as} \quad \beta(\alpha, \omega) \\
&Pr(X_{jk} | V_{1p}, V_{2q}) \quad \textit{distributed as} \quad \beta(\alpha_1, \omega_1).
\end{aligned}
$$

*Then,*

*$Pr(X_{il} | V_{1p}, V_{2q}, V_{3r})$ is distributed as $\beta(\alpha_2, \omega_2)$, where $\alpha_2 = \sum_{d \in \mathcal{D}_j} \frac{\alpha \alpha_1}{\sum_{f \in \mathcal{D}_1} \alpha_1}$ and $\omega_2 = \sum_{d \in \mathcal{D}_j} \frac{\omega \alpha_1}{\sum_{f \in \mathcal{D}_1} \alpha_1}$.*

**Proof.** The notation $\#\overline{X_{jk}}$ is the number of samples in which $X_j$ takes any value but $d_{j,k}$. The $\sum_{d \in \mathcal{D}_1} \alpha_1$ notation is the sum of the $\alpha$ statistics in the table for node $X_j$ that correspond to cells where $V_1$ takes any value, $X_j = d_{j,k}$, and $V_2 = d_{2,q}$. Recall that

$$
\begin{aligned}
\alpha_2 &= \#X_{il} \wedge V_{1p} \wedge V_{2q} \wedge V_{3r}, \text{ and} \\
\omega_2 &= \#\overline{X_{il}} \wedge V_{1p} \wedge V_{2q} \wedge V_{3r}.
\end{aligned}
$$

For $\alpha_2$:

$$
\begin{aligned}
\alpha_2 &= \#X_{il} \wedge V_{1p} \wedge V_{2q} \wedge V_{3r} \\
&= \textstyle\sum_{d \in \mathcal{D}_j} \#X_{jd} \wedge X_{il} \wedge V_{1p} \wedge V_{2q} \wedge V_{3r};
\end{aligned}
$$

then by the network assumption,

$$
\begin{aligned}
\alpha_2 &= \sum_{d \in \mathcal{D}_j} \frac{(\#X_{il} \wedge X_{jd} \wedge V_{2q} \wedge V_{3r})(\#V_{1p} \wedge X_{jd} \wedge V_{2q})}{(\#X_{jd} \wedge V_{2q})} \\
&= \sum_{d \in \mathcal{D}_j} \frac{\alpha \alpha_1}{\sum_{f \in \mathcal{D}_1} \alpha_1}.
\end{aligned}
$$

For $\omega_2$:

$$
\begin{aligned}
\omega_2 &= \#\overline{X_{il}} \wedge V_{1p} \wedge V_{2q} \wedge V_{3r} \\
&= \sum_{d \in \mathcal{D}_j} \frac{(\#X_{jd} \wedge \overline{X_{il}} \wedge V_{2q} \wedge V_{3r})(\#X_{jd} \wedge V_{1p} \wedge V_{2q})}{(\#X_{jd} \wedge V_{2q})} \\
&= \sum_{d \in \mathcal{D}_j} \frac{\omega \alpha_1}{\sum_{f \in \mathcal{D}_1} \alpha_1}.
\end{aligned}
$$

□

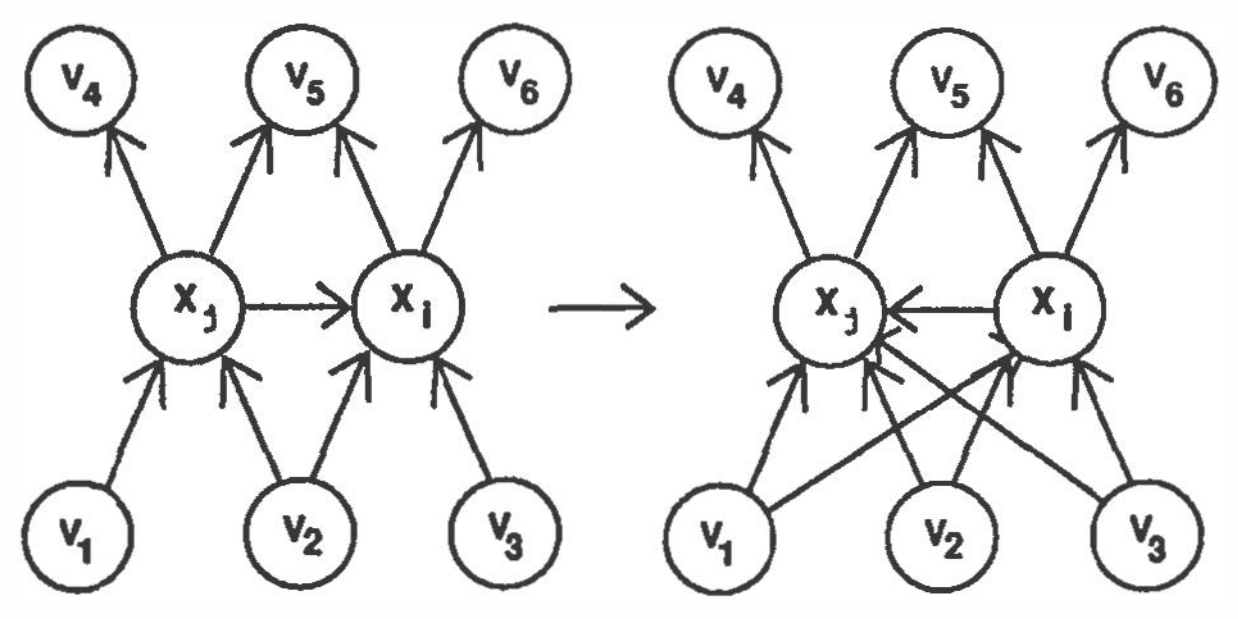

Figure 3: **Arc Reversal**
$V_1, V_2$ and $V_3$ are all defined as the Node Removal Figure. $V_4, V_5$ and $V_6$ are defined similarly, but represent sets of children of $X_i$ and $X_j$.

### Theorem 4.2 Arc Reversal

*Given:*
*The Network Assumption, and the network fragment as described in Figure 3, where $V_1$ represents the set of nodes which are parents of $X_j$ but not $X_i$; $V_2$ represents the set of parents common to both $X_i$ and $X_j$; and $V_3$ is the set of nodes which are parents of $X_i$ but not $X_j$. $V_4, V_5$ and $V_6$ are defined similarly, as children of $X_i$ and $X_j$. Also, given:*

$$\begin{array}{lll} Pr(X_{il}|X_{jk}, V_{2q}, V_{3r}) & \textit{distributed as} & \beta(\alpha, \omega) \\ Pr(X_{jk}|V_{1p}, V_{2q}) & \textit{distributed as} & \beta(\alpha_1, \omega_1). \end{array}$$

*Then,*

*$Pr(X_{il}|V_{1p}, V_{2q}, V_{3r})$ is distributed as $\beta(\alpha_2, \omega_2)$, where $\alpha_2 = \sum_{d\in\mathcal{D}_j} \frac{\alpha\alpha_1}{\sum_{f\in\mathcal{D}_1} \alpha_1}$ and $\omega_2 = \sum_{d\in\mathcal{D}_j} \frac{\omega\alpha_1}{\sum_{f\in\mathcal{D}_1} \alpha_1}$,*

*and*

*$Pr(X_{jk}|X_{il}, V_{1p}, V_{2q}, V_{3r})$ is distributed as $\beta(\alpha_3, \omega_3)$, where $\alpha_3 = \frac{\alpha\alpha_1}{\sum_{d\in\mathcal{D}_1} \alpha_1}$ and $\omega_3 = \alpha_2 - \alpha_3$.*

**Proof.** The proof is very similar to the proof of the Node Removal Theorem. The first portion of the theorem that finds $Pr(X_{il}|V_{1p}, V_{2q}, V_{3r})$ has already been proven in the Node Removal Theorem.

For $\alpha_3$:

$$\begin{aligned} \alpha_3 &= \#X_{jk} \wedge X_{il} \wedge V_{1p} \wedge V_{2q} \wedge V_{3r} \\ &= \frac{(\#X_{il}\wedge X_{jk}\wedge V_{2q}\wedge V_{3r})(\#V_{1p}\wedge X_{jk}\wedge V_{2q})}{(\#X_{jk}\wedge V_{2q})} \\ &= \frac{\alpha\alpha_1}{\sum_{d\in\mathcal{D}_1} \alpha_1}. \end{aligned}$$

For $\omega_3$:

$$\begin{aligned} \omega_3 &= \#\overline{X_{jk}} \wedge X_{il} \wedge V_{1p} \wedge V_{2q} \wedge V_{3r} \\ &= \#X_{il} \wedge V_{1p} \wedge V_{2q} \wedge V_{3r} - \\ &\quad \#X_{jk} \wedge X_{il} \wedge V_{1p} \wedge V_{2q} \wedge V_{3r} \\ &= \alpha_2 - \alpha_3. \end{aligned}$$

□

### Theorem 4.3 Node Merging

*Given:*
*The Network Assumption, and the network fragment as described in Figure 4, where $V_1$ represents the set of nodes which are parents of $X_j$ but not $X_i$; $V_2$ represents the set of parents common to both $X_i$ and $X_j$; and $V_3$ is the set of nodes which are parents of $X_i$ but not $X_j$. Also, given:*

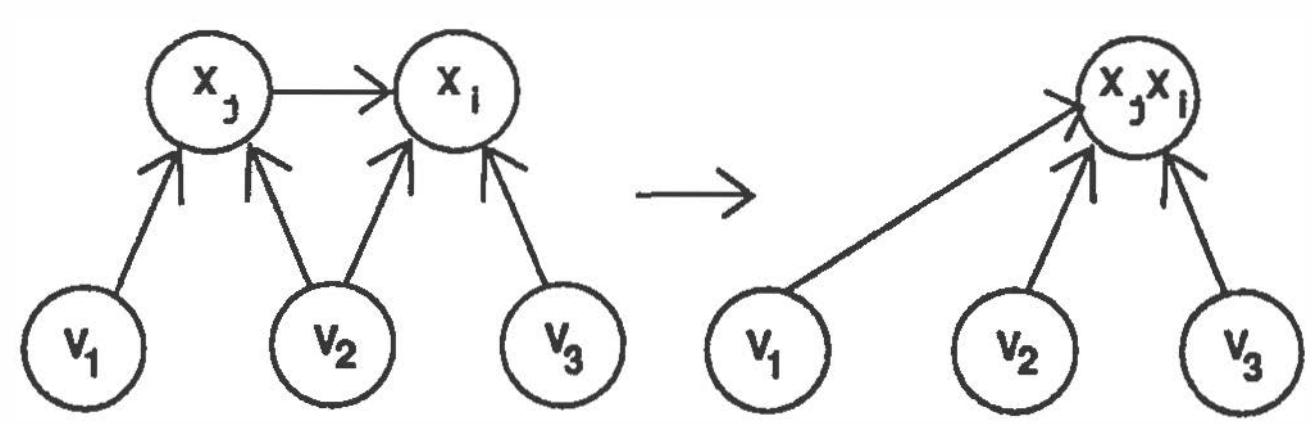

Figure 4: **Node Merging**

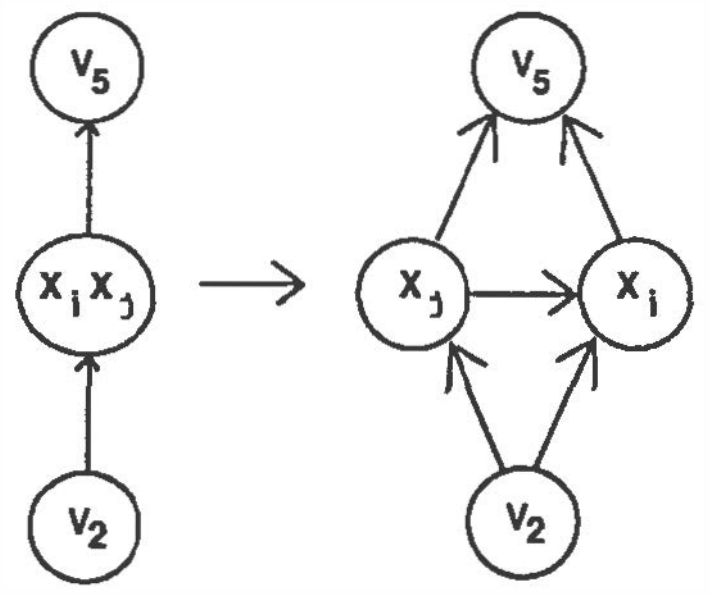

Figure 5: **Node Splitting**

$$\begin{array}{lll} Pr(X_{il}|X_{jk}, V_{2q}, V_{3,r}) & \textit{distributed as} & \beta(\alpha, \omega) \\ Pr(X_{jk}|V_{1p}, V_{2q}) & \textit{distributed as} & \beta(\alpha_1, \omega_1). \end{array}$$

*Then,*

*$Pr(X_{il}, X_{jk}|V_{1p}, V_{2q}, V_{3r})$ is distributed as $\beta(\alpha_2, \omega_2)$, where $\alpha_2 = \frac{\alpha\alpha_1}{\sum_{d\in\mathcal{D}_1} \alpha_1}$ and*
$\omega_2 = \frac{\alpha\alpha_1}{\sum_{f\in\mathcal{D}_1} \alpha_1} + \sum_{d\in\mathcal{D}_j} \frac{(\omega-\alpha)\alpha_1}{\sum_{f\in\mathcal{D}_1} \alpha_1}$.

**Proof.** The result for $\alpha_2$ was proven above in the Arc Reversal theorem.

For $\omega_2$:

$$\begin{aligned} \omega_2 &= \#\overline{X_{il} \wedge X_{jk}} \wedge V_{1p} \wedge V_{2q} \wedge V_{3r} \\ &= \#\overline{X_{il}} \wedge V_{1p} \wedge V_{2q} \wedge V_{3r} - \\ &\quad \#\overline{X_{jk}} \wedge X_{il} \wedge V_{1p} \wedge V_{2q} \wedge V_{3r}, \end{aligned}$$

and from the Arc Reversal Proof,

$$\begin{aligned} \omega_2 &= \sum_{d\in\mathcal{D}_j} \frac{\omega\alpha_1}{\sum_{f\in\mathcal{D}_1} \alpha_1} - \\ &\quad \left(\sum_{d\in\mathcal{D}_j} \frac{\alpha\alpha_1}{\sum_{f\in\mathcal{D}_1} \alpha_1} - \frac{\alpha\alpha_1}{\sum_{f\in\mathcal{D}_1} \alpha_1}\right) \\ &= \frac{\alpha\alpha_1}{\sum_{f\in\mathcal{D}_1} \alpha_1} + \sum_{d\in\mathcal{D}_j} \frac{(\omega-\alpha)\alpha_1}{\sum_{f\in\mathcal{D}_1} \alpha_1}. \end{aligned}$$

□

### Theorem 4.4 Node Splitting

*Given:*
*The Network Assumption, and the network fragment as described in Figure 5, where $V_2$ represents the set of parents common to both $X_i$ and $X_j$; and $V_5$ is the set of children for both nodes. Also, given:*

$$Pr(X_{il}, X_{jk}|V_{2q}) \quad \textit{distributed as} \quad \beta(\alpha, \omega).$$

*Then,*

*$Pr(X_{jk}|V_{2q})$ is distributed as $\beta(\alpha_1, \omega_1)$, where $\alpha_1 = \sum_{d \in \mathcal{D}_i} \alpha$, and $\omega_1 = \sum_{d \in \mathcal{D}_i} \sum_{f \in \mathcal{D}_j, f \neq k} \alpha$;*

*and*

*$Pr(X_{il}|X_{jk}, V_{2q})$ is distributed as $\beta(\alpha_2, \omega_2)$, where $\alpha_2 = \alpha$, and $\omega_2 = \sum_{d \in \mathcal{D}_i, d \neq l} \alpha$.*

**Proof.** Again the proof is very similar to the proof of the Node Removal Theorem.

For $\alpha_1$:

$$\begin{aligned} \alpha_1 &= \#X_{jk} \wedge V_{2q} \\ &= \textstyle\sum_{d \in \mathcal{D}_i} \alpha. \end{aligned}$$

For $\omega_1$:

$$\begin{aligned} \omega_1 &= \#\overline{X_{jk}} \wedge V_{2q} \\ &= \textstyle\sum_{d \in \mathcal{D}_i} \#\overline{X_{jk}} \wedge X_{id} \wedge V_{2q} \\ &= \textstyle\sum_{d \in \mathcal{D}_i} \sum_{f \in \mathcal{D}_j, f \neq k} \#X_{jf} \wedge X_{id} \wedge V_{2q} \\ &= \textstyle\sum_{d \in \mathcal{D}_i} \sum_{f \in \mathcal{D}_j, f \neq k} \alpha. \end{aligned}$$

For $\alpha_2$:

$$\begin{aligned} \alpha_2 &= \#X_{il} \wedge X_{jk} \wedge V_{2q} \\ &= \alpha. \end{aligned}$$

For $\omega_2$:

$$\begin{aligned} \omega_2 &= \#\overline{X_{il}} \wedge X_{jk} \wedge V_{2q} \\ &= \textstyle\sum_{d \in \mathcal{D}_i, d \neq l} \#X_{id} \wedge X_{jk} \wedge V_{2q} \\ &= \textstyle\sum_{d \in \mathcal{D}_i, d \neq l} \alpha. \end{aligned}$$

□

The final thing to show in this section is an outline of the proof that these transformations are sufficient for any inference in a belief network. Consider an inference in the form of Equation 1. Choose any source node (a node with no children) in the network, and label that the target. Choose a parent; if the parent shows up in the desired probability, use the Node Merging transformation to merge the parent with the target, otherwise use Node Removal to remove it. Do this until the only node left in the network is the target node. Then for every node in $\Psi$, perform the Node Splitting transformation to split off those nodes from the target. The end result is that the conditional probability table for the target will explicitly contain the inference distribution asked for in Equation 1. This approach would, of course, be horribly slow in practice.

## 5 EXAMPLE

In this section we work through a numerical example to get a better feeling for what these theorems mean, and how to apply them. The example network in Figure 6 is taken from Charniak (Charniak, 1991). It is a qualitative model used by a fictitious Mr. Smith to predict whether his family is out or not. The prediction is based on whether the outdoor lights are on, and whether the dog's barking can be heard. When the family leaves, they often put the dog out, and turn the outdoor light on. Also, when the dog is having bowel problems, he might be put out. An early indication of the dog being out is whether or not his barking can be heard.

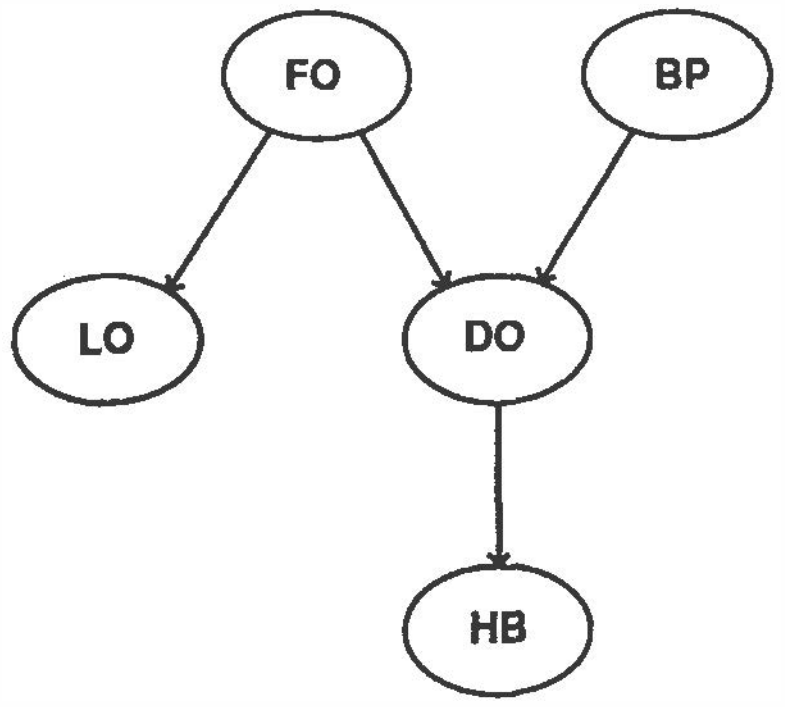

Figure 6: **The "family out" example**
Taken from Charniak, all variables are binary. FO = *fo* when the family is out, **fo** when not. Similarly, BP = *bp* when the dog has bowel problems, LO = *lo* when the outside lights are on, DO = *do* when the dog is out, and HB = *hb* when you can hear the dog barking.

| | *fo*, *bp* | *fo*, **bp** | **fo**, *bp* | **fo**, **bp** |
|---|---|---|---|---|
| *lo*, *do*, *hb* | 4 | 10 | 0 | 1 |
| *lo*, *do*, **hb** | 0 | 1 | 0 | 0 |
| *lo*, **do**, *hb* | 0 | 1 | 0 | 0 |
| *lo*, **do**, **hb** | 0 | 7 | 0 | 5 |
| **lo**, *do*, *hb* | 1 | 4 | 3 | 10 |
| **lo**, *do*, **hb** | 0 | 1 | 0 | 1 |
| **lo**, **do**, *hb* | 0 | 0 | 0 | 1 |
| **lo**, **do**, **hb** | 0 | 3 | 3 | 44 |

Table 1: **The Frequency distribution for FO,BP,LO,DO,HB**

Instead of attempting to construct the quantitative structure from memory of past experiences, Mr. Smith decides to start with an uninformative prior, and record his experiences daily (one can consider this an expensive sampling process for the more general task of sampling from a large scientific or business database). After 3 months, the sum total of his experience is represented by the frequency of each occurrence in the joint space, as shown in Table 1.

In general, tables representing the joint probability or frequency space will not be available, because they are usually far too large to represent outright. That is the raison d'être for belief networks: to efficiently represent the information in the joint space with a set of conditional probability tables, based on the independence assumptions characterized by the graphical layout (the Network Assumption). What is shown in Table 2 is the set of conditional probability tables that would be generated for the QBN representation of the data. For example, the entry for $Pr(hb|do) = \beta(34, 4)$ comes from the prior distribution of $\beta(1, 1)$ in each cell, plus 33 instances in which the dog was barking while outside, and 3 instances in which the dog was outside but not barking.

| *Pr*(FO) | |
|---|---|
| *fo* | β(33, 69) |
| **fo** | β(69, 33) |

| *Pr*(BP) | |
|---|---|
| *bp* | β(12, 90) |
| **bp** | β(90, 12) |

| *Pr*(HB\|DO) | | |
|---|---|---|
| | *do* | **do** |
| *hb* | β(34, 4) | β(3, 63) |
| **hb** | β(4, 34) | β(63, 3) |

| *Pr*(LO\|FO) | | |
|---|---|---|
| | *fo* | **fo** |
| *lo* | β(24, 10) | β(7, 63) |
| **lo** | β(10, 24) | β(63, 7) |

| *Pr*(DO\|FO,BP) | | | | |
|---|---|---|---|---|
| | *fo*, *bp* | *fo*, **bp** | **fo**,*bp* | **fo**, **bp** |
| *do* | β(6, 1) | β(17, 12) | β(4, 4) | β(13, 51) |
| **do** | β(1, 6) | β(12, 17) | β(4, 4) | β(51, 13) |

Table 2: **Conditional probability tables for QBN**
These tables are what would actually be stored, given the data as seen in Table 1.

For this example, we compute the $Pr(fo|lo, hb)$, and watch how the inference distribution is propagated throughout the transformation process. One possible set of transformations that can be performed on Figure 6 to get the desired distribution is shown in Figure 7. This transformation process proceeds as follows:

1. Arc Reversal: FO → LO. Following Theorem 4.2, the resulting distribution for $Pr$(FO= $fo$| LO=$lo$) = $\beta(24, 7)$. There are five other distributions (the rest of the table for $Pr$(FO|LO), and the table for $Pr$(LO)) that could be specified at this point on the basis of Theorem 4.2, but this is the only one of interest for this example.
2. Node Removal: BP. The distribution of $Pr$(DO=$do$| FO=$fo$) = $\beta(22, 12)$.
3. Arc Reversal: DO → HB. The distribution of $Pr$(DO=$do$|FO=$fo$,HB=$hb$) = $\beta$(20.25, 1.34). This is the first time the Network Assumption has cost accuracy. If we look at Table 1, we can see that the actual distribution based on the samples seen is $\beta(20, 2)$. The joint table, however, is not stored, so we must rely on the conditional probability tables for partial information, and the Network Assumption for the rest. In this case, an independence assumption was made that was not 100% supported by the data, and so the discrepancy. This is not too surprising, since it would be rare for a random sample of any size to completely support the independence conditions. Larger samples would, however, tend to provide stronger support.
4. Arc Reversal: FO → HB. The distribution of $Pr$(FO=$fo$| LO=$lo$,HB=$hb$) = $\beta(15.08, 2.36)$. This is the final result of the computation. Comparing this to the distribution we would get if Table 1 were available, we can see that the results are reasonable ($\beta(16, 2)$).

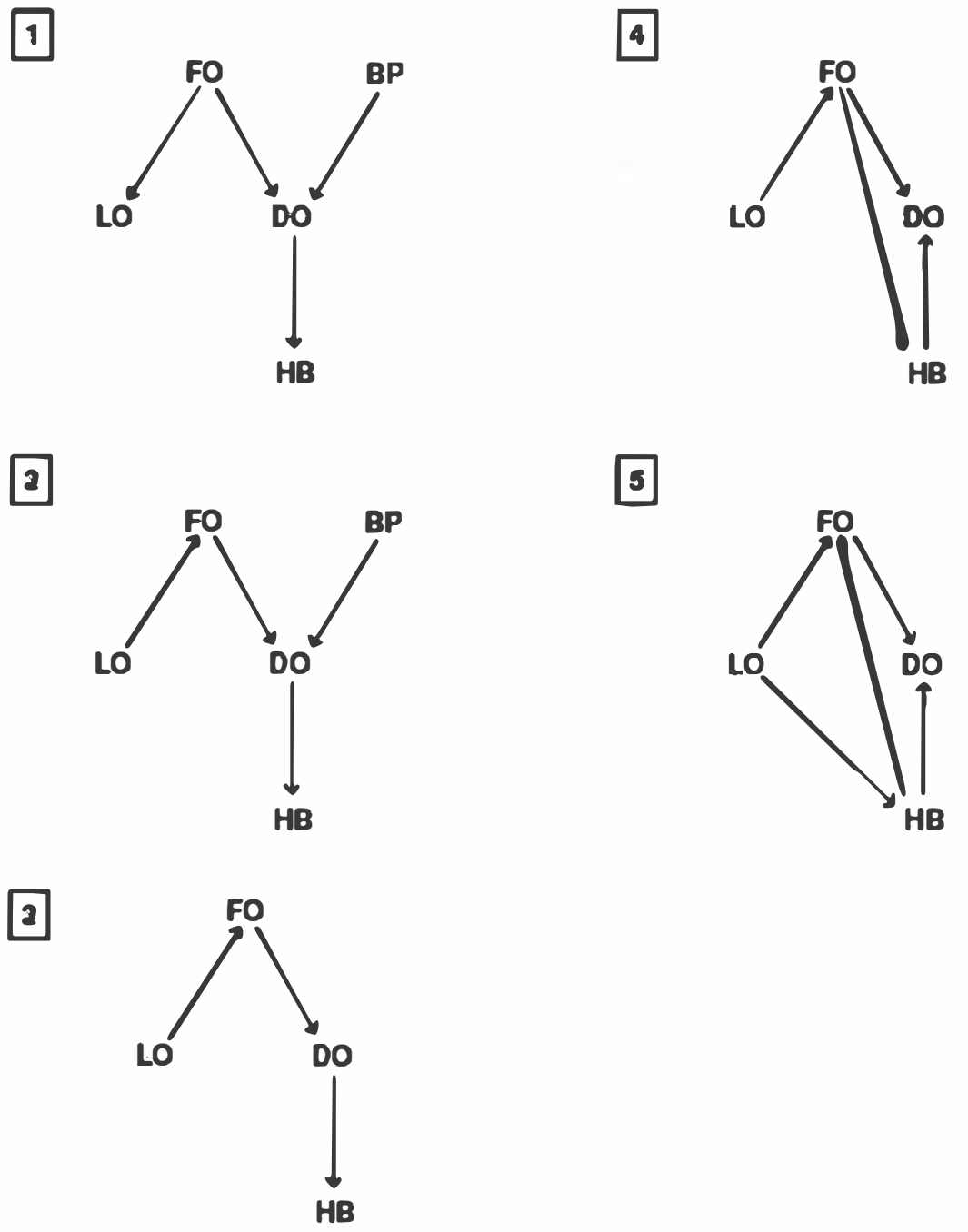

Figure 7: **Transformation steps to find the $Pr(fo|lo, hb)$**
The sequence of steps is labeled in each upper left hand corner.

It is widely accepted that as the inference chain grows, the error in the distributions being maintained will also grow. This statement is a bit too strong. As pointed out by Klieter (Klieter, 1992), there are two competing forces that will determine whether error or expertise is growing:

- As the inference chain grows, the error at each step is compounded, thus providing a force for increasing error.
- As the chain grows, certain types of inference (for example, marginalization) implicitly add samples from other parts of the network to the information currently at hand. Since the variance of the beta distribution is

$$
\begin{aligned}
var(\beta(\alpha,\omega)) &= \frac{\alpha\omega}{(\alpha+\omega)^2(\alpha+\omega+1)} \\
&< \frac{(\alpha+\omega)^2}{(\alpha+\omega)^2(\alpha+\omega+1)} \\
&= \frac{1}{\alpha+\omega+1},
\end{aligned}
$$

  the variance will tend to decrease linearly as samples are added to the system, thus providing a force for increasing expertise.

The magnitude of the first force depends on how the distributions are propagated. The larger the error in each step, the stronger this force. For the theory presented here, the source of the error is in the violation of the Network Assumption, in particular due to the sampled data not conforming to the independence claims made by the qualitative structure. As the size of the sampled data grows, this source of error will drop off fairly rapidly. It seems more reasonable to say that there exists a probabilistic *boundary* on the sample size (Musick and Russell, 1992). Below this boundary the error will grow as the inference chain grows, above it, expertise will grow instead.

## 6 CONCLUSION

In systems that learn belief networks by induction, there is a need to produce inference distributions rather than simple point probabilities in order to give some feeling for the degree confidence that can be placed in the result. It is impossible to produce distributions that accurately reflect all of the data seen without retaining the equivalent of the joint probability space for the network; there are too many unknowns, too few equations. It is necessary, then, to make assumptions to be capable of propagating distributions through inference. These assumptions have been strong in the past, compromising the integrity of the resulting distributions.

What this paper shows is that by using only the Network Assumption, the inference distributions can be propagated for any inference possible in a belief network, by specifying how to update the sufficient statistics on the beta distributions for four fundamental network transformations. Also, this paper has argued that under some types of inference, error will tend to decrease as the inference chain grows, as long as the belief network has been constructed from a sufficient number of instances.

For future work, the manipulations must be written in terms of realistic "fast" inference techniques like those explored in Lauritzen and Spiegelhalter (Lauritzen and Spiegelhalter, 1988; Lauritzen, 1992). Furthermore, learning the quantitative structure of a belief net does not end with the management of the beta distributions; we need to examine the possibility of learning good conditional probability tables with few samples. The quality of inference done in an induced network is dependent on the level to which the joint probability space has been explored. The fact of the matter is that for many domains, resources (the size of the database, the time available) will not be sufficient to adequately explore the joint probability space for a belief network.

## Acknowledgements

Thanks to Stuart Russell and Gary Ogasawara for many useful discussions. This material is based in part upon work supported by the National Science Foundation under Infrastructure Grant No. CDA-8722788, and the PYI Grant No. IRI-9058427 to Professor Stuart Russell.